\documentclass{article} 
\usepackage{iclr2027_conference,times}

\usepackage{amsmath,amsfonts,bm}

\def\eqref#1{equation~\ref{#1}}

\def\1{\bm{1}}

\DeclareMathAlphabet{\mathsfit}{\encodingdefault}{\sfdefault}{m}{sl}
\SetMathAlphabet{\mathsfit}{bold}{\encodingdefault}{\sfdefault}{bx}{n}

\usepackage{xcolor}
\usepackage{soul}
\definecolor{citeblue}{RGB}{0,90,180}
\definecolor{sentenceblue}{RGB}{225,239,255}
\DeclareRobustCommand{\bluehl}[1]{{\sethlcolor{sentenceblue}\hl{#1}}}
\usepackage{hyperref}
\hypersetup{colorlinks=true,citecolor=citeblue,linkcolor=red,urlcolor=citeblue}
\soulregister{\ref}{7}
\usepackage{url}
\usepackage{amsthm}
\newtheorem{proposition}{Proposition}
\newtheorem{corollary}{Corollary}
\usepackage{graphicx}
\usepackage{booktabs}
\usepackage{wrapfig}
\usepackage{wrapfig}
\usepackage{booktabs}
\usepackage{graphicx}
\usepackage{caption}
\usepackage{booktabs}
\usepackage{multirow}
\usepackage{colortbl}

\title{Exploring the Design Factors of Pixel-space Diffusion}

\author{Mang Ning  \\
Utrecht University \\
\And
Le Zhang \\
Mila - Quebec AI Institute \\
\AND
Albert Ali Salah \\
Utrecht University \\
\And
Itir Onal Ertugrul \\
Utrecht University \\
\And
Jintao Zhang \thanks {Corresponding author.} \\
Tsinghua University \\
}

\iclrfinalcopy 
\begin{document}

\maketitle
\lhead{} 

\begin{abstract}
   In this paper, we introduce EiT by exploring the key design factors of pixel-space diffusion from three perspectives: model architecture, representation, and frequency domain. Inspired by Denoising Autoencoders (DAE) which perform denoising through a bottleneck, we construct a U-shaped pixel diffusion transformer that performs explicit compression along the channel dimension. In terms of representation, we find that both pixel and latent diffusion transformers exhibit a three-stage representation dynamics across layers and timesteps: denoising/local feature expansion, basis lift-up, and posterior projection. Moreover, earlier denoising and a longer basis lift-up stage are associated with better generation, which EiT naturally achieves through its channel bottleneck and increased network depth. From the frequency perspective, we establish the equivalence between pixel-space and frequency-space diffusion when keeping the network operating in pixel space. On ImageNet $256\times256$ and $512\times512$, EiT consistently outperforms JiT (Just image Transformers) in generation quality. Specifically, our EiT achieves better FID than JiT: 1.61 vs. 1.85 at $256\times256$, and 1.73 vs. 1.99 at $512\times512$. Remarkably, EiT can be seamlessly applied to video pixel diffusion by simply extending the patch size to temporal dimension and outperforms video latent diffusion in FVD evaluated on Kinetics-600 (resolution 256), under comparable total network parameters. Code is available at \url{https://github.com/forever208/EiT}.
\end{abstract}

\section{Introduction}
\label{Section: Introduction}

Diffusion models \citep{ho2020denoising, song2020score} and flow matching \citep{lipman2022flow,liu2022flow} have rapidly advanced visual generation. For high-resolution synthesis, the dominant paradigm is latent diffusion \citep{rombach2022highresolution,peebles2023scalable,ma2024sit}: a variational autoencoder (VAE) \citep{kingma2013auto} first compresses images into a compact latent space, where a diffusion model is then trained. However, this two-stage pipeline is expensive, since the VAE requires careful design and costly training, and sampling requires running both the diffusion model and the decoder. Consequently, end-to-end pixel diffusion has attracted growing attention and demonstrated strong competitiveness \citep{hoogeboom2023simple,crowson2024scalable,ma2026deco,chen2026dip,yu2026pixeldit}. A representative example is JiT \citep{li2026back}, which shows that a plain diffusion transformer with large patches performs remarkably well under $x$-prediction. Nevertheless, pixel diffusion still lags behind state-of-the-art latent diffusion \citep{singh2026improved}.

In this paper, we explore the design factors of pixel diffusion from three perspectives: model architecture, representation, and frequency domain. First, we notice that $x$-prediction diffusion networks (at least in the low-noise regime) resemble Denoising Autoencoders (DAE), which are designed for representation learning \citep{vincent2008extracting}. Narrowing the hidden dimension forces DAE to keep only the dominant signal directions while discarding noise \citep{vincent2010stacked,baldi1989neural}. Motivated by DAE, we introduce compression into the pixel diffusion transformer, unifying representation compression and denoising in a single end-to-end network, EiT. Unlike prior U-shaped designs that merge tokens spatially \citep{crowson2024scalable,guo2026pixelu}, EiT compresses the channel dimension while keeping the token grid unchanged. On ImageNet, EiT significantly outperforms the plain diffusion transformer JiT and spatial compression transformer PixelU \citep{guo2026pixelu}. Second, by analyzing the effective rank and noise suppression, we reveal a three-stage representation dynamics in both pixel and latent diffusion transformers: denoising/local feature expansion, basis lift-up, and posterior projection. We interpret basis lift-up as expanding posterior-relevant information into a rich set of directions, from which the final layer can predict the target by linear projection. We further observe that earlier noise suppression and a longer stage of basis lift-up usually lead to better generation. EiT naturally falls into this regime: its channel bottleneck suppresses noise more effectively, and the saved computation affords a longer stage of basis lift-up. Finally, we prove that pixel-space and frequency-space diffusion are equivalent under orthonormal transforms when the network inputs and outputs remain in pixel space. Frequency-space diffusion nevertheless offers additional flexibility: reweighting the loss across frequency bands can improve generation quality without extra parameters or FLOPs. Overall, our contributions are:

\begin{itemize}
\item We propose EiT, a pixel diffusion transformer with channel compression, which outperforms JiT and PixelU under comparable budgets and also benefits latent diffusion. Importantly, EiT can be easily extended to video pixel diffusion and outperforms video latent diffusion on Kinetics-600.
\item We uncover a three-stage representation dynamics shared by pixel and latent diffusion, where earlier noise suppression and longer basis lift-up are associated with better generation.
\item We prove the equivalence of pixel-space and frequency-space diffusion, and show that frequency-band loss reweighting further improves generation at no extra cost.
\end{itemize}

\section{Related Work}
\label{Section: Related Work}

\paragraph{Pixel diffusion.}
Early pixel diffusion models make high-resolution training feasible through shifted noise schedules, sigmoid loss weighting, and hierarchical UNet or hourglass Transformer backbones \citep{hoogeboom2023simple,crowson2024scalable,hoogeboom2025simpler}. A promising work is JiT \citep{li2026back}: under the manifold assumption, clean images lie on a low-dimensional manifold, so predicting the clean image ($x$-prediction) lets a plain large-patch Transformer model high-dimensional pixels even with limited capacity, where $\epsilon$- and $v$-prediction fail catastrophically. Concurrently, DeCo, DiP, and PixelDiT \citep{ma2026deco,chen2026dip,yu2026pixeldit} share a frequency-decoupled design: a DiT on large patches models low-frequency semantics, while a lightweight pixel-level module restores high-frequency details. Other VAE-free methods exploit the frequency structure of images more explicitly, including diffusion in the DCT space \citep{ning2024dctdiff}, coarse-to-fine resolution cascades \citep{chen2025pixelflow}, and band-specific noise schedules in the wavelet domain \citep{pavasovic2026wait}. Most related to our architecture are HDiT \citep{crowson2024scalable} and PixelU \citep{guo2026pixelu}, which compress representations along the spatial dimension by merging neighboring tokens. In contrast, we explore compression along the channel dimension while keeping the token grid intact.

\paragraph{Diffusion representation.}
Research connecting diffusion models and representations follows two directions. The first treats diffusion models as representation learners. Diffusion Autoencoders pair a semantic encoder with a diffusion decoder to obtain decodable latent codes \citep{preechakul2022diffusion}, and DiffMAE casts masked image modeling as conditional diffusion for visual pretraining \citep{wei2023diffusion}. Pretrained diffusion features also transfer to understanding tasks: DIFT extracts intermediate activations for zero-shot correspondence \citep{tang2023emergent}, and RepFusion distills time-dependent features into recognition networks \citep{yang2023diffusion}. The second direction uses representations to improve generation. REPA aligns noisy hidden states with clean-image features from pretrained encoders \citep{yu2024representation}, while SRA aligns its shallow, high-noise features with its own deeper, low-noise ones \citep{jiang2025no}. RAE instead performs diffusion directly in the latent space of frozen representation encoders \citep{zheng2026diffusion}, and RiT shows that a vanilla diffusion transformer with $x$-prediction suffices in this space \citep{zhang2026rit}. However, these works either evaluate diffusion features by downstream utility or exploit representations as supervision and latent spaces. In this paper, we dive deep into the representation dynamics in diffusion transformers and reveal how they relate to the posterior target learning.

\section{Analysis and Method}
\label{Section: Analysis and Method}

\subsection{Background: Diffusion and Flow}
\label{Subsection: Background: Diffusion and Flow}

Diffusion models and Flow matching models can be formulated from the perspective of
ODEs (Ordinary Differential Equations) \citep{song2020score, lipman2022flow}. For simplicity, we consider the flow matching formulation for the discussion in this paper. Given the data sample $ \pmb{x} \sim p_{data}$ and Gaussian noise $ \pmb{\epsilon} \sim \mathcal{N}(0, \pmb{I})$, a noisy sample is constructed by interpolation $\pmb{x}_t = (1-t) \pmb{x} + t \pmb{\epsilon}$ under the linear schedule \citep{liu2022flow} in the time interval $t \in [0, 1]$. The conditional velocity is defined as $\pmb{v} = \frac{d \pmb{x}_t}{dt} = \pmb{\epsilon} - \pmb{x}$ and is learned by a model $\pmb{v}_\theta$. The flow matching objective is

\noindent
\begin{equation}
\label{eq: flow-matching-objective}
\mathcal{L}_{\mathrm{flow}}(\theta) = \mathbb{E}_{ \pmb{x}, \pmb{\epsilon}, t } \left\| \pmb{v}_\theta(\pmb{x}_t, t) - \pmb{v} \right\|_2^2.
\end{equation}

Given the learned flow velocity $\pmb{v}_\theta$, the sampling is obtained by solving the ODE $\frac{d \pmb{x}_t}{dt} = \pmb{v}_\theta(\pmb{x}_t, t)$ starting from $\pmb{x}_1 \sim \mathcal{N}(0, \pmb{I})$ and ending at $t=0$. Following \citet{li2026back}, we adopt the $x$-prediction network to obtain the velocity function $\pmb{v}_\theta(\pmb{x}_t, t)$.

\subsection{U-shaped Architecture and Channel Compression}
\label{Subsection: U-shaped Architecture and Channel Compression}

\begin{figure}[ht]
   \vskip -0.1in
   \begin{center}
   \centerline{\includegraphics[width=1.0\columnwidth]{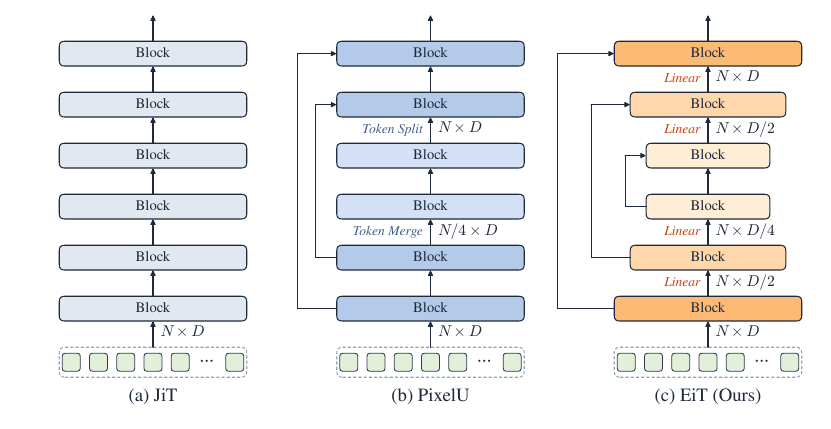}}
   \caption{Comparison of pixel diffusion Transformer architectures.
   JiT keeps both the token count $N$ and hidden dimension $D$ fixed.
   PixelU constructs a spatial bottleneck by merging and splitting tokens.
   EiT preserves the full token grid and instead forms a channel bottleneck
   through linear projection; long
   skips connect mirrored encoder and decoder blocks.}
   \label{fig: arch_compare}
   \end{center}
   \vskip -0.2in
\end{figure}

Latent diffusion decomposes image generation into two stages: a VAE first maps an image to a compact latent space, and a diffusion model then learns the score or velocity in this compressed space \citep{rombach2022highresolution}. In contrast, a diffusion model must learn its velocity field directly in the high-dimensional pixel space, and a standard isotropic Transformer processes representations of the same size throughout the network. Inspired by DAE that performs denoising through a bottleneck, we incorporate compression into the pixel diffusion Transformer. The resulting EiT model performs diffusion and representation compression jointly in one end-to-end network. Existing U-shape pixel diffusion Transformers compress representations along the spatial dimension. For example, HDiT repeatedly merges neighboring tokens \citep{crowson2024scalable}, while PixelU uses one spatial downsampling stage \citep{guo2026pixelu}.  Unlike these token merge methods, we explore another direction: compressing the \textbf{channel dimension} while keeping the token number unchanged (see Figure~\ref{fig: arch_compare}). In the experiments section, we will show that this channel-compression strategy yields a better generation quality than compressing the spatial dimension.

To make our channel compression a generic and simple method, we make the minimum changes to the original diffusion transformers \citep{ma2024sit,li2026back} by keeping the head dimension $d$ fixed in self-attention and the number of tokens unchanged. \textbf{The channel compression is achieved by varying only the number of attention heads} $h_\ell$, thus the channel dimension is $D_\ell = h_\ell d$ at layer $\ell$. Since the number of tokens and dimension of self-attention are unchanged, no effort is needed to modify the positional embedding (e.g., 2D RoPE \citep{su2024roformer}). Moreover, the encoder and decoder are symmetric and blocks are uniformly distributed across stages. Therefore, the instantiation of EiT is easy and flexible. For instance, we can instantiate a 4-stage schedule $4$-$3$-$2$-$1$, corresponding to encoder channel dimensions $(D,3D/4,2D/4,D/4)$, and the decoder reverses the selected schedule. Following U-ViT \citep{bao2023all} and PixelU \citep{guo2026pixelu}, we connect mirrored encoder and decoder blocks with long skips. As in U-Net \citep{ronneberger2015u}, these connections provide a path for high-frequency details to bypass the compression bottleneck. Our ablation in Appendix~\ref{Appendix: Effect of long-skip connection in JiT and EiT} confirms their importance for EiT, while showing no benefit for the plain, constant-width JiT.


Channel compression is particularly efficient for image generation. It is known that much of the computation in a Transformer \citep{vaswani2017attention} is spent in the feed-forward network (FFN), whose cost scales as $\mathcal{O}(ND^2)$, rather than in attention with cost $\mathcal{O}(N^2D)$ in the small $N$ regime. Reducing the channel dimension from $D$ to $\alpha D$ therefore reduces the parameters and FLOPs of FFN to approximately $\alpha^2$ of their original values. Thus, the saved computation can be used to increase the depth under matched parameter and FLOP budgets with JiT. Furthermore, our analysis in Section~\ref{Subsection: Diffusion Representation Dynamics} suggests two benefits of EiT architecture: the channel bottleneck suppresses noise better than JiT, while the increased depth provides a longer basis lift-up stage before the final posterior projection.

\subsection{Diffusion Representation Dynamics}
\label{Subsection: Diffusion Representation Dynamics}

In the $x$-prediction network, the diffusion transformer is trained to directly predict the posterior mean $\mathbb{E}[\pmb{x}\mid\pmb{x}_t]$ which does not contain random noise at any timestep $t$. Therefore, the network should learn representations to perform denoising. However, is denoising the only behaviour we should expect from all layers and timesteps? To understand the internal representation dynamics in diffusion transformer, we apply the \textbf{Spatial Participation Ratio} (similar to the entropy effective rank) to characterize the dimensionality of the spatial representation, and use Noise Fraction to infer the noise capacity in the representation
.  Specifically, let
$\pmb{H}^{\ell,t}(\pmb{x}_t, t)\in\mathbb{R}^{N\times D}$ denote the
layer-$\ell$ activation whose rows are the $N$ spatial tokens.  We subtract the mean spatial token and then compute the singular values
$\{\sigma_i\}$. Defining the normalized spectral
energy as $p_i=\sigma_i^2/\sum_j\sigma_j^2$, the Spatial Participation Ratio (SPR) is

\noindent
\begin{equation}
R_{\mathrm{PR}}
=\frac{1}{\sum_i p_i^2}
=\frac{\left(\sum_i\sigma_i^2\right)^2}
{\sum_i\sigma_i^4}.
\end{equation}

\begin{figure}[ht]
   \vskip -0.1in
   \begin{center}
   \centerline{\includegraphics[width=1.0\columnwidth]{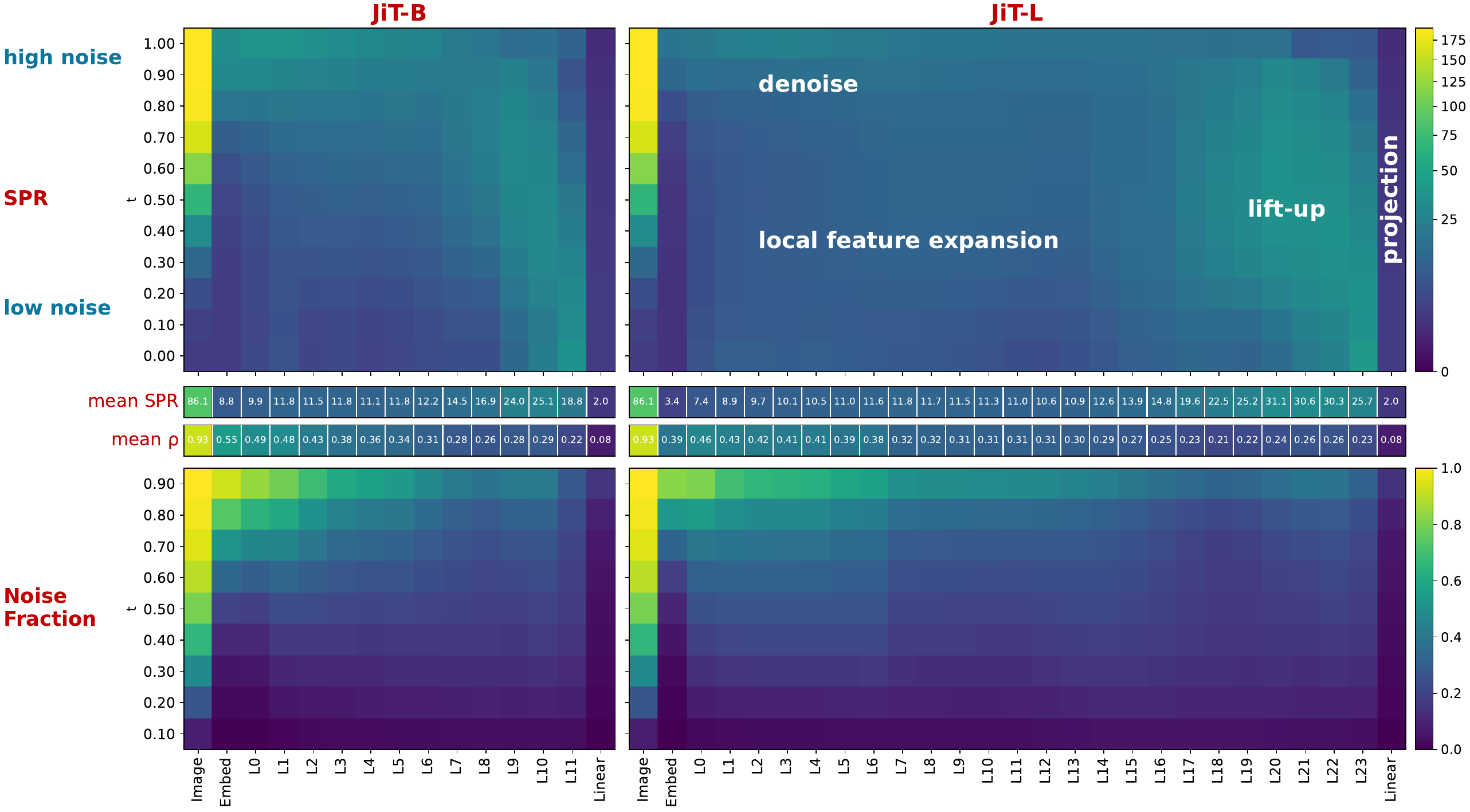}}
   \caption{SPR and Noise Fraction of JiT-B (left) and JiT-L (right) on ImageNet $256{\times}256$.
   Columns follow the network: noisy input (Image), patch embedding (Embed), each Transformer block (L0, L1, \dots), and the final linear layer (Linear); larger $t$ means higher noise. Middle: mean SPR over $t\in[0.1,0.9]$ and mean Noise Fraction $\rho$ over $t\in[0.5,0.9]$.
   }
   \label{fig: JIT PR Noise}
   \end{center}
   \vskip -0.2in
\end{figure}

Like entropy effective rank \citep{roy2007effective},  SPR summarizes a singular-value spectrum as the effective number of active directions.  
We choose SPR because it emphasizes dominant modes and is less affected by a weak spectral tail than entropy effective rank. In our analysis, we compute the SPR for each layer and timestep of JiT models on ImageNet 256 and plot the results in the first row of Figure \ref{fig: JIT PR Noise}. However, SPR is also highly sensitive to noise: Gaussian noise flattens the singular-value spectrum and can substantially increase SPR. Thus, a high SPR can indicate either much noise or a signal distributed across many dimensions. To distinguish these cases, we introduce an intervention-based \textbf{Noise Fraction} as a proxy for noise content in the representation.  Let
$\phi_{\ell,t}(\pmb{x},\pmb{\epsilon})$
denote the per-token normalized form of the representation $\pmb{H}^{\ell,t}(\pmb{x},\pmb{\epsilon})$.  We measure the noise-induced and signal-induced variations as
\begin{align}
N_{\ell,t}
&=\mathbb{E}_{\pmb{x},\pmb{\epsilon}_a,\pmb{\epsilon}_b}
\left\|
\phi_{\ell,t}(\pmb{x},\pmb{\epsilon}_a)
-\phi_{\ell,t}(\pmb{x},\pmb{\epsilon}_b)
\right\|_F^2, \\
S_{\ell,t}
&=\mathbb{E}_{\pmb{\epsilon},\pmb{x}_a,\pmb{x}_b}
\left\|
\phi_{\ell,t}(\pmb{x}_a,\pmb{\epsilon})
-\phi_{\ell,t}(\pmb{x}_b,\pmb{\epsilon})
\right\|_F^2,
\end{align}
where $N_{\ell,t}$ varies the noise while keeping the image fixed, whereas
$S_{\ell,t}$ varies images while sharing the same noise.  We then define
\begin{equation}
\rho_{\ell,t}
=\frac{N_{\ell,t}}{N_{\ell,t}+S_{\ell,t}}.
\end{equation}
A large $\rho_{\ell,t}$ indicates a noise-sensitive representation, while a
small value indicates that its variation is dominated by the clean signal.
Consequently, high SPR together with a high Noise Fraction suggests a noise-dominated representation, whereas high SPR with a low Noise Fraction is consistent with an expansion of image-related feature directions. The SPR and Noise Fraction results are plotted in Figure \ref{fig: JIT PR Noise}. To better summarize the behaviour of each layer, we also report the mean SPR and mean Noise Fraction of each layer in the middle rows of Figure \ref{fig: JIT PR Noise}, where SPR is averaged over $t\in[0.1,0.9]$ and Noise Fraction over the high-noise range $t\in[0.5,0.9]$, in which denoising is demanding.

Through the lens of SPR and Noise Fraction, we find that \bluehl{diffusion transformer exhibits a 3-stage representation dynamics across layers and timesteps: denoising/local feature expansion - basis lift-up - posterior projection}. We summarize the model behavior as:

\begin{itemize}
\item \textbf{denoising in shallow layers \& high-noise $\pmb{x}_t$}: the Noise Fraction gradually reduces as the network depth increases
\item \textbf{local feature expansion in shallow layers \& low-noise $\pmb{x}_t$}: the SPR increases after the FFN layer inside the transformer block while the Noise Fraction remains low.
\item \textbf{basis lift-up in deep layers}: the SPR dramatically increases while the Noise Fraction is very low.
\item \textbf{Posterior projection in the final linear layer}: low SPR aligning with the posterior target.
\end{itemize}

Clearly, denoising is not the only behaviour in the diffusion transformer. \textbf{But why does lift-up emerge?} Our interpretation is that the posterior mean
$m_t(\pmb{x}_t)=\mathbb{E}[\pmb{x}\mid\pmb{x}_t]$ is generally a complex nonlinear function, whereas the final layer must predict it through a
linear projection. We therefore interpret the sharp SPR increase in deep layers as basis lift-up: the nonlinear backbone expands posterior-relevant information into a richer, overcomplete set of basis directions in which the posterior becomes easier to approximate with a linear projection. Supporting this interpretation, our linear-probing experiments show that the clean $\pmb{x}$ is already linearly decodable from the early layers of lift-up, before the final projection. This interpretation is consistent with dimensional expansion followed by task-aligned compression in deep networks \citep{recanatesi2019dimensionality}, and echoes the
feature-lifting principle behind universal approximation \citep{cybenko1989approximation} and kernel methods \citep{rahimi2007random}. In Appendix \ref{Appendix: Effect of supressing the lift-up}, we also empirically verify that supressing the basis lift-up would degrade the generation quality in pixel diffusion. Interestingly, the depth of Transformer provides a plausible contribution to this lift-up effect: under residual updates
$\pmb{h}_{\ell+1}=\pmb{h}_{\ell}+F_{\ell}(\pmb{h}_{\ell},t)$, each block can add or refine nonlinear features, yielding an iterative enrichment of the computational basis before the final projection. Consequently, a deeper Transformer has more opportunities to accumulate such target-aligned directions, which may explain why diffusion transformers usually scale well with depth \citep{peebles2023scalable}.

\begin{wraptable}{r}{0.5\textwidth}
   \vspace{-1.0\baselineskip}
   \centering
   \captionsetup{skip=2pt}
   \caption{Comparison of representation dynamics and generation quality.}
   \label{tab:good_diffusion_representation}
   \resizebox{\linewidth}{!}{
      \begin{tabular}{@{}lllll@{}}
         \toprule
         Model       & Depth & Denoising & Lift-up & FID \\ \midrule
         JiT-B       & 12    &     -      &   -      &    3.59       \\
         JiT-L       & 24    &     earlier      &   more layers      &    2.41       \\ \midrule
         SiT-XL      & 28    &     -      &   -      &    2.54       \\
         SiT-XL-REPA & 28    &     earlier      &   more layers      &    1.68       \\ \bottomrule
         \end{tabular}
   }
   \vspace{-0.0\baselineskip}
   \end{wraptable}

\begin{figure}[ht]
   \vskip -0.1in
   \begin{center}
   \centerline{\includegraphics[width=1.0\columnwidth]{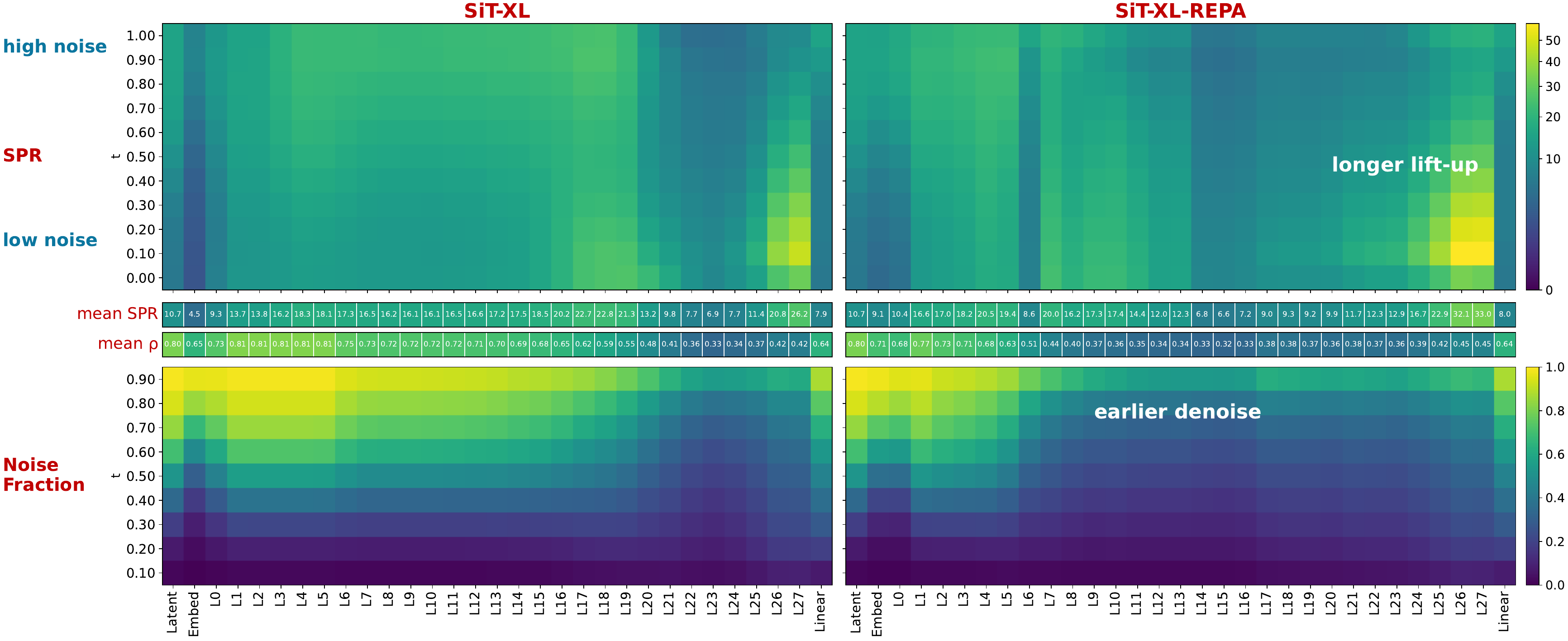}}
   \caption{SPR and Noise Fraction of SiT-XL (left) and SiT-XL-REPA (right) on ImageNet $256{\times}256$, with the same layout as Figure~\ref{fig: JIT PR Noise} (Latent denotes the noisy VAE latent input).
   Compared with SiT-XL, SiT-XL-REPA denoises at earlier layers and sustains a high SPR over more deep layers.}
   \label{fig: SIT PR Noise}
   \end{center}
   \vskip -0.2in
\end{figure}

   We further find that these representation dynamics are not exclusive to pixel-space diffusion but also arise in latent-space diffusion. Figure~\ref{fig: SIT PR Noise} shows the SPR and Noise Fraction of SiT-XL \citep{ma2024sit} and SiT-XL-REPA \citep{yu2024representation}, both evaluated on ImageNet $256{\times}256$. Although both models are trained to predict the noise-containing target $\mathbb{E}[\pmb{\epsilon}-\pmb{x}\mid\pmb{x}_t]$, they still exhibit the same three-stage dynamics. The two comparisons in Table~\ref{tab:good_diffusion_representation} suggest two potential characteristics of effective diffusion representations: \textbf{earlier noise suppression and a longer basis lift-up stage}. Specifically, increasing JiT's depth from 12 to 24 layers and adding REPA to SiT-XL both produce earlier denoising and a longer lift-up stage, together with improved FID. Interestingly, we find that EiT suppresses noise more effectively than JiT (Appendix~\ref{Appendix: Diffusion Representation Dynamics}) because the U-shape bottleneck encourages the network to discard nuisance noise while retaining the principal component of the image signal \citep{baldi1989neural}, which resembles DAE \citep{vincent2008extracting}. 
   In Appendix~\ref{Appendix: Diffusion Representation Dynamics}, we also demonstrate that models with longer lift-up stage usually achieve better posterior prediction and FID.
   However, whether a long lift-up stage is necessary for good generation quality remains an open question.

\subsection {Equivalence of Frequency-space Diffusion and Pixel-space Diffusion}
\label{Subsection: Equivalence of Frequency-space Diffusion and Pixel-space Diffusion}

After discussing the model architecture and representation dynamics, we now explore the third direction of pixel diffusion: frequency domain. Previous studies have shown that performing image generative modeling purely in the Discrete Cosine Transform (DCT) space \citep{ning2024dctdiff} or Discrete Wavelet Transform (DWT) space \citep{pavasovic2026wait} can achieve superior generation quality than pixel-space diffusion under heuristic exploitation of frequencies and noise schedules. To reveal the role of frequency diffusion, we formally show that, \bluehl{if the network inputs and outputs are both kept in pixel space, frequency-space and pixel-space diffusion are equivalent under an orthonormal transformation (Proposition \ref{prop:orthonormal-equivalence}).}

\begin{proposition}[\textbf{Equivalence under orthonormal transform and coupled randomness}]
\label{prop:orthonormal-equivalence}
Let $\pmb{W}\in\mathbb{R}^{n\times n}$ satisfy
$\pmb{W}^\top\pmb{W}=\pmb{W}\pmb{W}^\top=\pmb{I}$, and explicitly couple the pixel-space and frequency-space noises by $\pmb{\epsilon}_z:=\pmb{W}\pmb{\epsilon}$ with
$\pmb{\epsilon}\sim\mathcal{N}(0,\pmb{I})$. Consider
\[
\pmb{x}_t=(1-t)\pmb{x}+t\pmb{\epsilon},\qquad
\pmb{z}_t=(1-t)\pmb{W}\pmb{x}+t\pmb{\epsilon}_z,
\]
where both spaces use the same linear schedule. Define
$\widetilde{\pmb{x}}_t=\pmb{W}^\top\pmb{z}_t$ and the corresponding conditional
velocity targets
\[
\pmb{v}=\pmb{\epsilon}-\pmb{x},\qquad
\pmb{v}_z=\pmb{\epsilon}_z-\pmb{W}\pmb{x}.
\]
For the same pixel-space predictor architecture $\pmb{v}_\theta$, define the per-sample losses
\[
L_{\mathrm{pix}}(\theta)
=\|\pmb{v}_\theta(\pmb{x}_t,t)-\pmb{v}\|_2^2,\qquad
L_{\mathrm{freq}}(\theta)
=\|\pmb{W}\pmb{v}_\theta(\widetilde{\pmb{x}}_t,t)-\pmb{v}_z\|_2^2.
\]
For every coupled training sample and every $\theta$,
\[
\widetilde{\pmb{x}}_t=\pmb{x}_t,\qquad
\pmb{v}_z=\pmb{W}\pmb{v},\qquad
L_{\mathrm{freq}}(\theta)=L_{\mathrm{pix}}(\theta),
\]
and the two losses have identical gradients wherever differentiable. Hence
their population objectives
$\mathcal{L}_{\mathrm{pix}}(\theta)$
and
$\mathcal{L}_{\mathrm{freq}}(\theta)$
are equal.

Suppose further that the two runs start from identical parameter--optimizer states and couple all stochastic training variables at every step, then pixel-space and frequency-space training follow the same optimization trajectory:
\[
\boxed{\theta_k^{\mathrm{freq}}=\theta_k^{\mathrm{pix}}
\quad\text{for every training step }k.}
\]
Consequently, the two models obtain exactly the same final parameters.
\end{proposition}

\begin{corollary}[\textbf{Equivalence in expected loss under uncoupled randomness}]
\label{cor:uncoupled-training-equivalence}
Under the same architecture and optimizer configuration, suppose that all stochastic variables in the two runs are sampled independently according to the same laws. Then, at every training step $k$, the expected population losses over repeated training runs are equal:
\[
\mathbb{E}_{\mathrm{train}}\!\left[
\mathcal{L}_{\mathrm{freq}}(\theta_k^{\mathrm{freq}})
\right]
=
\mathbb{E}_{\mathrm{train}}\!\left[
\mathcal{L}_{\mathrm{pix}}(\theta_k^{\mathrm{pix}})
\right].
\]
\end{corollary}

A complete proof of Proposition~\ref{prop:orthonormal-equivalence} and Corollary~\ref{cor:uncoupled-training-equivalence} is given in Appendix~\ref{Appendix:proof-orthonormal-equivalence}.

\begin{figure}[ht]
   \vskip 0.0in
   \begin{center}
   \centerline{\includegraphics[width=0.9\columnwidth]{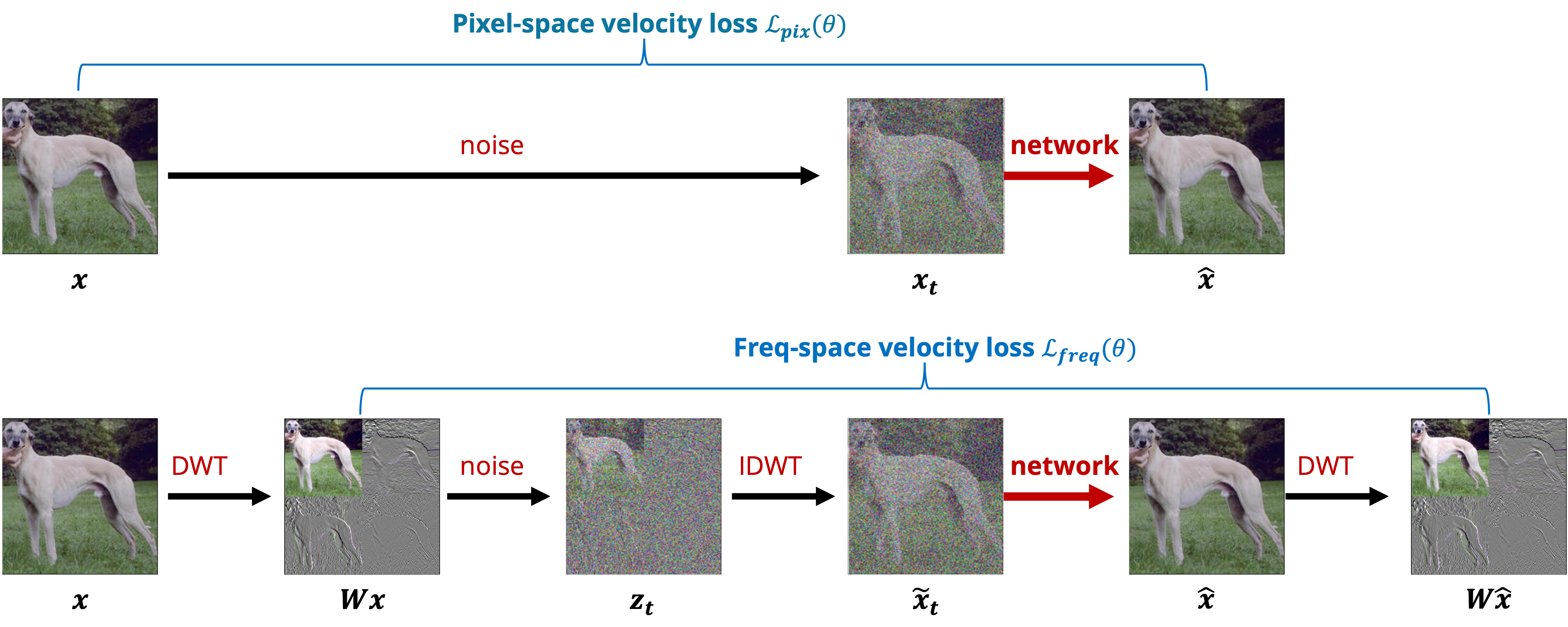}}
   \caption{Pixel-space diffusion training (first row) and Frequency-space diffusion training (second row) on ImageNet 256. The orthonormal matrix $\pmb{W}$ can be either DWT or DCT.
   }
   \label{fig: DWT Flow Matching}
   \end{center}
   \vskip -0.2in
\end{figure}

Intuitively, Proposition~\ref{prop:orthonormal-equivalence} states that pixel and frequency diffusion training follow exactly the same optimization trajectory when all training randomness is shared. Corollary~\ref{cor:uncoupled-training-equivalence} relaxes this coupling requirement so that individual trajectories may differ, but their expected losses over repeated runs remain equal. The normalized DCT and the Haar DWT are classical orthonormal transforms \citep{ahmed1974discrete, mallat1989theory} and are special cases of $\pmb{W}$, so the two pipelines in Figure~\ref{fig: DWT Flow Matching} are expected to achieve very close performance. We verify this equivalence by training JiT-B (pixel) and JiT-B (DWT) using the two pipelines in Figure~\ref{fig: DWT Flow Matching}. On ImageNet 256$\times$256, JiT-B (DWT) achieves FID 3.60, almost the same as FID of 3.59 obtained by JiT-B (Pixel). We also observe that \textbf{keeping the network operating in pixel space is crucial for this equivalence}, as directly feeding DWT coefficients to the network degrades the FID from 3.60 to 3.71 (see Appendix~\ref{Appendix: Empirical Verification of Equivalence}).

After clarifying the equivalence, \textbf{what are the advantages of frequency diffusion over pixel diffusion?} DCTdiff \citep{ning2024dctdiff} shows that frequency reweighting can exploit the varying sensitivity of human vision to different frequencies to improve spectral generative modeling, a distinctive advantage of frequency-space modeling over pixel-space modeling. DeCo \citep{ma2026deco} shares the same spirit by weighting frequency-domain losses using a DCT quantization table. Similarly, we can apply frequency reweighting to the frequency diffusion loss $\mathcal{L}_{\mathrm{freq}}(\theta)$. For example, using level-1 Haar DWT yields four frequency bands: LL, HH, HL, LH \citep{daubechies1992ten}. We can assign higher weight to the LL component of $\mathcal{L}_{\mathrm{freq}}(\theta)$ to enhance the generation of global structure or improve the sample quality of fine details by highlighting the HH/HL/LH components in $\mathcal{L}_{\mathrm{freq}}(\theta)$. Let $\pmb{r}_\theta=\pmb{W}\pmb{v}_\theta(\widetilde{\pmb{x}}_t,t)-\pmb{v}_z$ denote the DWT velocity residual, and let $\pmb{r}_{\theta,b}$ denote its coefficients in frequency band $b$. We propose the DWT loss reweighting as:

\noindent
\begin{equation}
\label{eq: dwt-velocity-loss}
\mathcal{L}_{\mathrm{DWT}}(\theta)
=\mathbb{E}_{\pmb{x},\pmb{\epsilon}_z,t}\!\left[
\lambda\|\pmb{r}_{\theta,\mathrm{LL}}\|_2^2
+\gamma\!\left(
\|\pmb{r}_{\theta,\mathrm{HH}}\|_2^2
+\|\pmb{r}_{\theta,\mathrm{HL}}\|_2^2
+\|\pmb{r}_{\theta,\mathrm{LH}}\|_2^2
\right)\right].
\end{equation}

Empirically, we observe that enhancing the low-frequency component, by setting $\lambda>1$ and $\gamma=1$, is beneficial on the ImageNet dataset. Thus we also propose an alternative low-frequency enhancement method for pixel diffusion: Low-Resolution Regularization (LRR). Its key idea is to construct an auxiliary flow-matching path at a lower spatial resolution, strengthening the modeling of low-frequency content. Let $\pmb{x}_{\mathrm{low}}$ denote the image obtained by downsampling
$\pmb{x}$. We independently draw $\pmb{\epsilon}_{\mathrm{low}}\sim\mathcal{N}(0,\pmb{I}_{\mathrm{low}})$ and construct $\pmb{x}_{t,\mathrm{low}} =(1-t)\pmb{x}_{\mathrm{low}}+t\pmb{\epsilon}_{\mathrm{low}}$ and $\pmb{v}_{\mathrm{low}}=\pmb{\epsilon}_{\mathrm{low}}-\pmb{x}_{\mathrm{low}}$. We augment the original velocity objective in Equation~\ref{eq: flow-matching-objective} with this low-resolution loss:

\noindent
\begin{equation}
\label{eq: low-resolution-regularization}
\mathcal{L}_{\mathrm{LRR}}(\theta)
=\mathbb{E}_{\pmb{x},\pmb{\epsilon},
\pmb{\epsilon}_{\mathrm{low}},t}\!\left[
\|\pmb{v}_\theta(\pmb{x}_t,t)-\pmb{v}\|_2^2
+\eta
\|\pmb{v}_\theta(\pmb{x}_{t,\mathrm{low}},t)
-\pmb{v}_{\mathrm{low}}\|_2^2
\right],
\end{equation}

Note that the network now performs twice the forward pass during training and $\pmb{x}_{\mathrm{low}}$ yields the additional $1/4$ patch tokens. Thus, LRR brings extra training cost while DWT loss reweighting does not.

\section{Experimental Results}
\label{Section: Experimental Results}

Considering the architecture simplicity, we build our model EiT from JiT for image generation tasks and compare our models against JiT and PixelU. To ensure fairness, we always use the same patch size and training recipe for EiT and the base model (unless otherwise noted). The full training parameters and model details are provided in the Appendix \ref{Appendix: Training Parameters and Model Details}. We report the generation results by Fréchet Inception Distance (FID) \citep{heusel2017gans} and Inception Score (IS) \citep{salimans2016improved} on ImageNet 256$\times$256 and 512$\times$512 \citep{russakovsky2015imagenet}. We leave all ablation studies to the Appendix \ref{Appendix: Ablation Studies}. Regariding the video generation tasks, we benchmark EiT and the latent diffusion model \citep{song2025history} on Kinetics-600 \citep{kay2017kinetics}.

\subsection{Results of EiT and EiT (DWT)}
\label{subection: results of channel compression - EiT}

\begin{wrapfigure}{r}{0.5\textwidth}
   \vspace{-1.0\baselineskip}
     \includegraphics[width=0.5\textwidth]{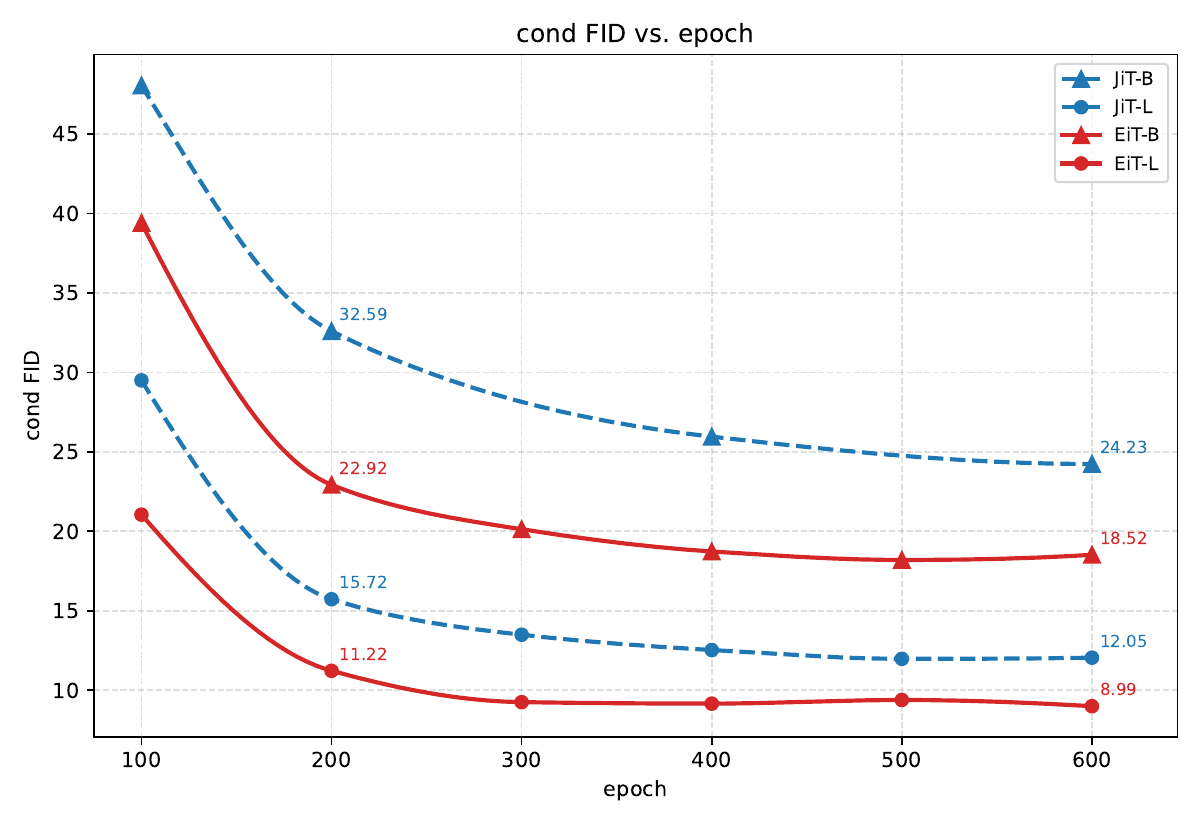}
     \captionsetup{skip=2pt}
     \caption{Conditional FID of Pixel Diffusion Models on ImageNet 256$\times$256.}
     \label{fig: conditional FID}
     \vspace{-0.4\baselineskip}
   \end{wrapfigure}

\paragraph{Results of EiT.}
We first evaluate the performance of EiT (channel compression) and compare it with JiT and PixelU (spatial compression) without REPA loss. In detail, we train all models for 600 epochs following the training parameters recommended by JiT \citep{li2026back} and split the experiments into two groups according to the model size: base and large. The generation quality is then evaluated by FID-50k and IS using 50-step Heun sampler \citep{karras2022elucidating}. For a detailed comparison, we perform conditional generation and classifier-free guidance generation (CFG) \citep{ho2022classifier} separately. As summarized in Table~\ref{tab: Generation comparison on ImageNet 256.}, EiT consistently achieves the best FID among models of comparable size on ImageNet 256$\times$256. Notably, despite using fewer parameters and less computation, EiT-L attains an FID of 2.22, still outperforming JiT-L (2.41) and PixelU-L (2.23). Conditional generation usually provides a more direct measure of the quality of the learned score function. Figure~\ref{fig: conditional FID} suggests that, after only 200 epochs, both EiT-B and EiT-L already outperform their JiT counterparts trained for 600 epochs in conditional FID. To further isolate the contribution of channel compression from network depth, we conduct a controlled ablation in which EiT-L and JiT-L both use 24 layers and have comparable parameters. Results in Appendix \ref{Appendix
: Contribution of Channel Compression in EiT} identify that channel compression is a key contributor to EiT's performance gains. Finally, JiT and EiT demonstrate similar wall-clock time efficiency during training: 328s per epoch for JiT-B and 332s per epoch for EiT-B.

\begin{table}[ht]
   \vskip -0.0in
   \captionsetup{skip=2pt}
   \caption{
   Generation Comparison of Pixel Diffusion Models on ImageNet 256$\times$256.}
   \label{tab: Generation comparison on ImageNet 256.}
   \begin{center}
      \begin{tabular}{@{}lllllll@{}}
         \toprule
         \multirow{2}{*}{Models} & \multirow{2}{*}{Depth} & \multirow{2}{*}{Hidden dim} & \multirow{2}{*}{Parameters} & \multirow{2}{*}{GFLOPs} & \multicolumn{2}{c}{CFG} \\ \cmidrule(lr){6-7}
         & & & & & FID $\downarrow$ & IS $\uparrow$ \\ \midrule
         JiT-B & 12 & 768 & 131M & 25.2 & 3.59 & 262.6 \\

         PixelU-B & 12 & 768 & 158M & 22.4 & 3.53 & 269.8 \\
         
         EiT-B & 24 & $\leq$768 & 130M & 24.3 & 3.10 & 275.3 \\
         
         EiT-B (DWT) & 24 & $\leq$768 & 130M & 24.3 & \textbf{3.02} & \textbf{277.2} \\ \midrule
         
         JiT-L & 24 & 1024 & 459M & 87.9 & 2.41 & 290.1 \\
         
         PixelU-L & 24 & 1024 & 516M & 75.3 & 2.23 & 291.6 \\
         
         EiT-L & 30 & $\leq$1024 & 363M & 67.1 & 2.22 & 303.3 \\
         
         EiT-L & 30 & $\leq$1152 & 460M & 84.4 & 2.06 & 300.2 \\
         
         EiT-L (DWT) & 30 & $\leq$1152 & 460M & 84.4 & \textbf{2.03} & \textbf{312.1} \\ \bottomrule
         \end{tabular}
   \end{center}
   \vskip -0.0in
\end{table}

\paragraph{Results of EiT (DWT).}
In Section~\ref{Subsection: Equivalence of Frequency-space Diffusion and Pixel-space Diffusion}, we established the equivalence between frequency-space and pixel-space diffusion and showed that the frequency-space formulation enables flexible frequency-band loss reweighting at negligible additional cost. To validate the effectiveness of DWT and its loss reweighting, we train EiT (DWT) using Equation~\ref{eq: dwt-velocity-loss} with $\lambda=1.2$ and $\gamma=1$ (namely highlighting the training of low-frequency components). As shown in Tables~\ref{tab: Generation comparison on ImageNet 256.}, EiT (DWT) consistently improves generation quality on ImageNet 256$\times$256 without increasing the parameter count or model GFLOPs. We also find that Low-Resolution Regularization (LRR) yields similar improvements for pixel-space diffusion (Appendix~\ref{Appendix: Loss Reweighting of DWT and Low-Resolution Regularization}), but incurs approximately 28\% additional training GFLOPs. These results demonstrate that DWT provides an effective and computationally efficient realization of frequency diffusion.

\subsection{Scaling EiT to Huge Size}
\label{subsection: Scaling EiT to Huge Size}

To validate the scaling effect of our EiT, we scale both EiT and JiT to 1B parameters (huge size) and train both models from scratch on ImageNet 256$\times$256 and ImageNet 512$\times$512. Then we compare EiT-H against other pixel diffusion models: Simple Diffusion \citep{hoogeboom2023simple}, PixelFlow \citep{chen2025pixelflow}, PixelNerd \citep{wang2026pixnerd}, DiP \citep{chen2026dip}, DeCo \citep{ma2026deco} and PixelDiT \citep{yu2026pixeldit} . As shown in Table \ref{tab: Large-scale generation comparison on ImageNet 256.}, EiT-H improves FID from 1.85 to 1.78 over JiT-H while using fewer GFLOPs, and achieves the highest IS of 339.0 among the compared models. Adding REPA, EiT-H achieves FID 1.61, matching PixelDiT, and the second-highest IS of 312.9. On ImageNet 512$\times$512, both JiT and EiT adopt a patch size 32, so that the number of image tokens maintain at 256. Results in Table~\ref{tab: Large-scale generation comparison on ImageNet 512.} show that EiT-H improves FID from 1.99 to 1.85 over JiT-H and achieves the highest IS of 361.6. With REPA, EiT-H further reduces FID to 1.73, outperforming all compared models. These results demonstrate EiT’s scalability and competitiveness against a broader range of pixel-space diffusion methods.

\begin{table}[ht]
   \vskip -0.0in
   \captionsetup{skip=2pt}
   \caption{
   Generation Comparison of Large-Scale Pixel Diffusion Models on ImageNet 256$\times$256.}
   \label{tab: Large-scale generation comparison on ImageNet 256.}
   \begin{center}
      \begin{tabular}{@{}lllllll@{}}
         \toprule
         \multirow{2}{*}{Models} & \multirow{2}{*}{\shortstack[l]{Training\\epochs}} & \multirow{2}{*}{REPA} & \multirow{2}{*}{Parameters} & \multirow{2}{*}{GFLOPs} & \multicolumn{2}{c}{CFG} \\ \cmidrule(lr){6-7}
         & & & & & FID $\downarrow$ & IS $\uparrow$ \\ \midrule
         Simple Diffusion & --- & $\times$ & 2B & 555 & 2.44 & 256.3 \\
         PixelFlow-XL & 320 & $\times$ & 677M & 2909 & 1.98 & 282.1 \\
         DiP-XL & 320 & $\checkmark$ & 631M & 128 & 1.90 & 285.4 \\
         DeCo-XL & 600 & $\checkmark$ & 682M & 123 & 1.69 & 304.0 \\
         PixelDiT-XL & 320 & $\checkmark$ & 797M & 156 & \textbf{1.61} & 292.7 \\
         JiT-H & 600 & $\times$ & 1050M & 200 & 1.85 & 308.6 \\
         JiT-G & 600 & $\times$ & 2B & 383 & 1.82 & 292.6 \\
         \rowcolor{sentenceblue}
         EiT-H & 600 & $\times$ & 1074M & 194 & 1.78 & \textbf{339.0} \\
         \rowcolor{sentenceblue}
         EiT-H (REPA) & 600 & $\checkmark$ & 1074M & 194 & \textbf{1.61} & 312.9 \\ \bottomrule
      \end{tabular}
   \end{center}
   \vskip -0.0in
\end{table}

\begin{table}[ht]
   \vskip -0.0in
   \captionsetup{skip=2pt}
   \caption{
   Generation Comparison of Large-Scale Pixel Diffusion Models on ImageNet 512$\times$512.}
   \label{tab: Large-scale generation comparison on ImageNet 512.}
   \begin{center}
      \begin{tabular}{@{}lllllll@{}}
         \toprule
         \multirow{2}{*}{Models} & \multirow{2}{*}{\shortstack[l]{Training\\epochs}} & \multirow{2}{*}{REPA} & \multirow{2}{*}{Parameters} & \multirow{2}{*}{GFLOPs} & \multicolumn{2}{c}{CFG} \\ \cmidrule(lr){6-7}
         & & & & & FID $\downarrow$ & IS $\uparrow$ \\ \midrule
         Simple Diffusion & --- & $\times$ & 2B & 555 & 3.02 & 248.7 \\
         PixelNerd-XL & --- & $\checkmark$ & 701M & 584 & 2.84 & 245.6 \\
         DiP-XL & --- & $\checkmark$ & 631M & 180 & 2.31 & 291.7 \\
         DeCo-XL & 340 & $\checkmark$ & 682M & 540 & 2.22 & 290.0 \\
         PixelDiT-XL & 850 & $\checkmark$ & 797M & 676 & 1.81 & 278.6 \\
         JiT-H & 600 & $\times$ & 1054M & 201 & 1.99 & 315.1 \\
         \rowcolor{sentenceblue}
         EiT-H & 600 & $\times$ & 1079M & 195 & 1.85 & \textbf{361.6} \\
         \rowcolor{sentenceblue}
         EiT-H (REPA) & 600 & $\checkmark$ & 1079M & 195 & \textbf{1.73} & 326.7 \\ \bottomrule
      \end{tabular}
   \end{center}
   \vskip -0.0in
\end{table}

\subsection{Extending EiT to Pixel-Space Video Generation}
\label{subsection: Extending EiT to Pixel-Space Video Generation}

In addition to image generation, we find that it is easy to extend our EiT architecture to the video generation task and train the pixel video diffusion model in an end-to-end manner. To have a fair comparison, we adopt the latent diffusion transformer DFoT \citep{song2025history} as the baseline and train both EiT and DFoT on the Kinetics-600 (256$\times$256) \citep{kay2017kinetics} for 40 epochs using the same training recipe (learning rate, batch size, etc.). In the settings of DFoT, 17 frames of each video are compressed into the latent space with 4$\times$ temporal downsampling and 16$\times$ spatial downsampling, corresponding to 1280 patch tokens. These video tokens are later trained by diffusion forcing \citep{chen2024diffusion}. During evaluation, the diffusion model receives 5 clean frames as history to perform conditional generation to synthesize the next 12 frames. 50-step DDIM sampler \citep{song2020denoising} is applied in sampling without CFG. The generation quality is evaluated by FVD \citep{unterthiner2018towards} and IS using 50k test samples. In the patch design of EiT, we merge 16$\times$16$\times$3$\times$4 pixels into a single token, leading to the same 1280 patch tokens as in DFoT. As shown in Table~\ref{tab: Generation Comparison of Video Diffusion Models on Kinetics-600.}, EiT achieves better FVD and IS than DFoT under comparable total parameters, demonstrating the effectiveness of our EiT architecture for video generation.

\begin{table}[ht]
   \vskip -0.0in
   \captionsetup{skip=2pt}
   \caption{Generation Comparison of Video Diffusion Models on Kinetics-600.}
   \label{tab: Generation Comparison of Video Diffusion Models on Kinetics-600.}
   \begin{center}
      \begin{tabular}{@{}lllllllll@{}}
         \toprule
         \multirow{2}{*}{Models} & \multicolumn{2}{c}{Diffusion} & \multicolumn{2}{c}{Encoder} & \multicolumn{2}{c}{Decoder} & \multirow{2}{*}{FVD $\downarrow$} & \multirow{2}{*}{IS $\uparrow$} \\ \cmidrule(lr){2-3} \cmidrule(lr){4-5} \cmidrule(lr){6-7}
         & \# para & GFLOPs & \# para & GFLOPs & \# para & GFLOPs & & \\ \midrule
         DFoT & 435M & 622 & 68M & 2170 & 108M & 10127 & 11.94 & 30.5 \\
         \rowcolor{sentenceblue}
         EiT & 653M & 958 & 0 & 0 & 0 & 0 & \textbf{10.75} & \textbf{31.3} \\ \bottomrule
      \end{tabular}
   \end{center}
   \vskip -0.0in
\end{table}

\subsection{Extending EiT to Latent-Space Diffusion}
\label{Subsection: Extend to latent diffusion}

\begin{table}[ht]
   \centering
   \captionsetup{skip=2pt}
   \caption{Generation Comparison of Latent Diffusion Models on ImageNet 256$\times$256.}
   \label{tab: Generation Comparison of Latent Diffusion Models on ImageNet 256.}
   {\normalsize
      \begin{tabular}{@{}llllll@{}}
         \toprule
         \multirow{2}{*}{Models} & \multirow{2}{*}{Parameters} & \multirow{2}{*}{GFLOPs} & \multirow{2}{*}{Epochs} & \multicolumn{2}{c}{FID} \\ \cmidrule(l){5-6} 
          &  &  &  & cond & CFG \\ \midrule
         SiT-B & 130M & 23.0 & 80 & 39.81 & 6.39 \\
         EiT-B & 129M & 22.3 & 80 & \textbf{29.17} & \textbf{4.99} \\ \midrule
         SiT-B-REPA & 138M & 24.9 & 80 & 24.55 & 3.39 \\
         EiT-B-REPA & 136M & 24.1 & 80 & \textbf{19.18} & \textbf{3.09} \\ \midrule
         SiT-XL-REPA & 683M & 120.7 & 200 & 7.40 & 1.68 \\
         EiT-XL-REPA & 685M & 117.2 & 200 & \textbf{7.13} & \textbf{1.64} \\ \bottomrule
      \end{tabular}
   }
\end{table}

To assess the generality of channel compression, we extend EiT to latent-space diffusion by incorporating its channel-compression architecture into SiT \citep{ma2024sit}. We evaluate the resulting models both with and without REPA \citep{yu2024representation} on ImageNet 256$\times$256, training all Base models for 80 epochs and all XL models for 200 epochs. As shown in Table~\ref{tab: Generation Comparison of Latent Diffusion Models on ImageNet 256.}, EiT consistently outperforms its SiT counterpart under comparable parameter and computation budgets. Without REPA, EiT-B reduces CFG FID from 6.39 to 4.99. The gains remain consistent when combined with REPA: EiT-B-REPA improves the FID scores from 3.39 to 3.09. These results demonstrate that channel compression is effective not only for pixel-space diffusion but also for latent-space diffusion.

\section{Conclusion}
\label{Section: Conclusion}

In this paper, we explored the design factors of pixel-space diffusion from three perspectives. For model architecture, we proposed EiT, which introduces channel compression into the pixel diffusion transformer and outperforms JiT and PixelU under comparable budgets, while also benefiting latent diffusion. Importantly, EiT is easy to extend to pixel-space video generation. For representation, we revealed a three-stage dynamics of denoising/local feature expansion, basis lift-up, and posterior projection shared by pixel and latent diffusion, and found that earlier noise suppression and a longer lift-up stage are associated with better generation. For the frequency domain, we proved the equivalence between pixel-space and frequency-space diffusion provided that the network inputs and outputs remain in pixel space. A limitation of this work is that our findings do not establish whether a longer basis lift-up stage is necessary for high-quality generation. We hope these observations encourage further theoretical and empirical research in the community.

\bibliography{iclr2027_conference}
\bibliographystyle{iclr2027_conference}

\newpage
\appendix

\section{Effect of suppressing the lift-up}
\label{Appendix: Effect of supressing the lift-up}

To verify whether basis lift-up benefits generation quality, we train a variant, JiT-B (SPR), which explicitly suppresses lift-up by adding an SPR penalty to the training loss. Specifically, the SPR of any layer exceeding 16 is penalized with a loss weight of 0.1. Both JiT-B (SPR) and JiT-B are trained for 600 epochs on ImageNet 256$\times$256. Suppressing the lift-up degrades generation quality, increasing the FID from 3.59 to 3.79 (with CFG), suggesting that basis lift-up is beneficial for pixel diffusion.

\section{Diffusion Representation Dynamics}
\label{Appendix: Diffusion Representation Dynamics}

\begin{figure}[ht]
   \vskip -0.1in
   \begin{center}
   \centerline{\includegraphics[width=1.0\columnwidth]{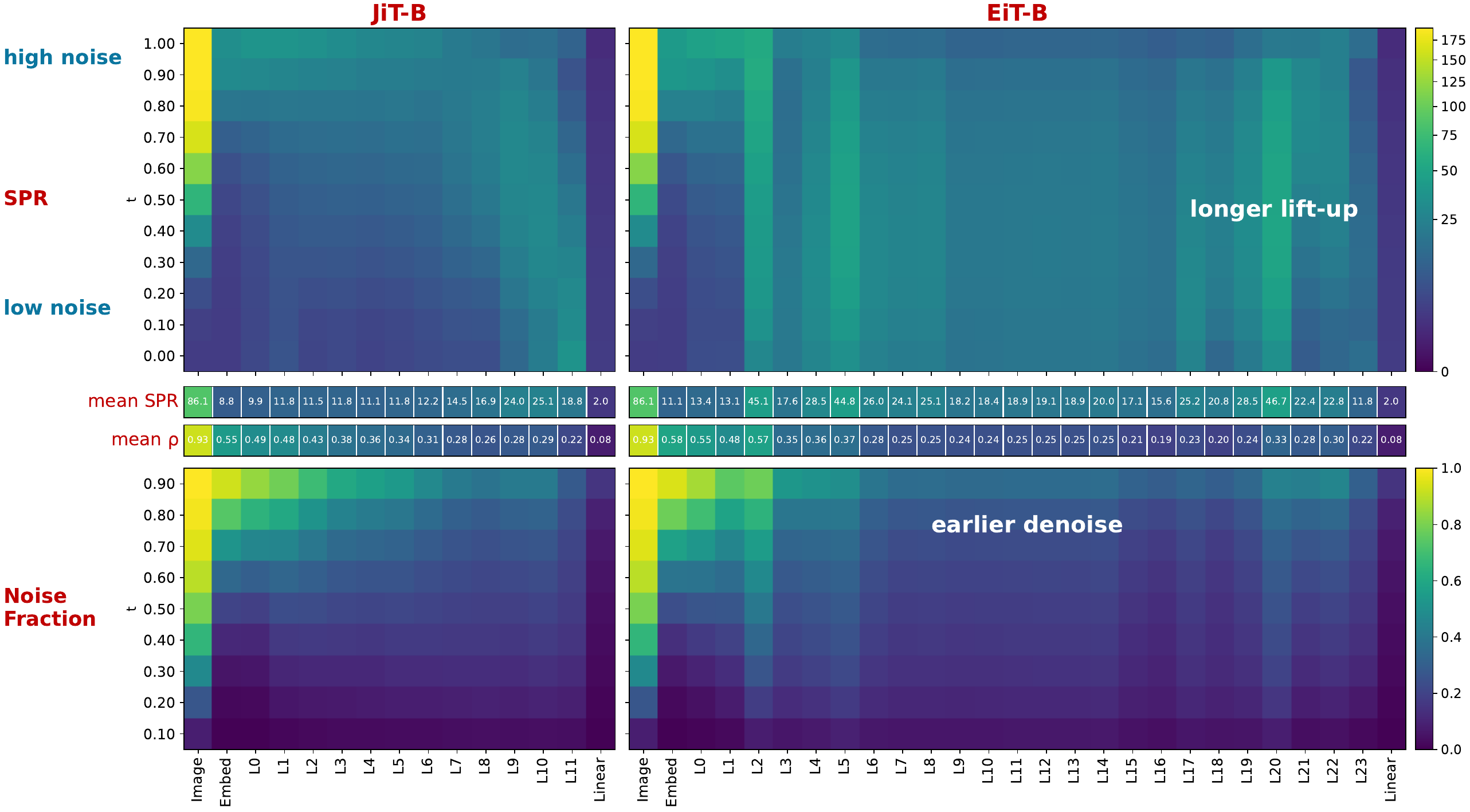}}
   \caption{SPR and Noise Fraction of JiT-B (12 blocks, left) and EiT-B (24 blocks, u-type $4$-$3$-$2$-$1$, right) on ImageNet $256{\times}256$, with the same layout as Figure~\ref{fig: JIT PR Noise}.
   Compared with JiT-B, EiT-B reduces the Noise Fraction at an earlier relative depth (earlier denoising) and maintains a high SPR over more deep blocks (longer lift-up).}
   \label{fig: EIT PR Noise}
   \end{center}
   \vskip -0.2in
\end{figure}

Figure~\ref{fig: EIT PR Noise} compares the SPR and Noise Fraction of JiT-B (12 blocks) and EiT-B (24 blocks, u-type $4$-$3$-$2$-$1$) on ImageNet 256$\times$256, following the protocol in Section~\ref{Subsection: Diffusion Representation Dynamics}. The mean rows average SPR over $t\in[0.1,0.9]$ and Noise Fraction over the high-noise range $t\in[0.5,0.9]$, where denoising is most demanding. EiT-B exhibits the same three-stage dynamics as JiT-B, with two notable differences.

\paragraph{Earlier and stronger noise suppression.} 
EiT-B reduces the mean Noise Fraction to 0.28 at L6, about one third of its depth, whereas JiT-B reaches this level at L7, about two thirds of its depth. The minimum mean Noise Fraction across Transformer blocks is 0.19 at L16 for EiT-B, versus 0.22 at L11 for JiT-B.

\paragraph{Longer basis lift-up.} 
In the deep layers, EiT-B sustains a mean SPR above 20 over six consecutive blocks (L17--L22), compared with two (L9--L10) in JiT-B, indicating a longer basis lift-up stage.

\paragraph{Longer basis lift-up associated with better posterior prediction and FID.} 
We apply posterior linear probes to the final-layer representations of pretrained models to assess their accuracy in predicting the posterior target. We focus on the low-noise interval $t\in[0.1,0.5]$, where we approximate the posterior target with the clean image $\pmb{x}$. We use 800/100/100 training/validation/test images per class and report test pixel MSE averaged equally over four classes and timesteps. Along with Posterior MSE, we add the Noise Fraction at the middle layer and the number of layers of lift-up (mean SPR above 20) in Table~\ref{tab:representation_posterior_fid}. 
Results show that a longer basis lift-up stage is associated with lower posterior MSE and FID within each model scale. For example, EiT-B has 6 lift-up layers versus 2 in JiT-B, with lower posterior MSE ($9.61\times10^{-3}$ vs.\ $9.65\times10^{-3}$) and FID (3.10 vs.\ 3.59). The same pattern holds for Large and Huge models (7 vs.\ 6 and 8 vs.\ 6 lift-up layers, respectively).

\begin{table}[ht]
   \centering
   \captionsetup{skip=2pt}
   \caption{Comparison of representation dynamics, posterior prediction error, and generation quality for JiT and EiT on ImageNet 256$\times$256.}
   \label{tab:representation_posterior_fid}
   \begin{tabular}{@{}lcccc@{}}
      \toprule
      Models & \shortstack{Noise Fraction} & Layers of lift-up & Posterior MSE $\downarrow$ & FID (CFG) \\ \midrule
      JiT-B & 0.34 & 2 & $9.65\times10^{-3}$ & 3.59 \\
      EiT-B & \textbf{0.25} & \textbf{6} & $\mathbf{9.61\times10^{-3}}$ & \textbf{3.10} \\ \midrule
      JiT-L & 0.31 & 6 & $9.45\times10^{-3}$ & 2.41 \\
      EiT-L & \textbf{0.24} & \textbf{7} & $\mathbf{9.42\times10^{-3}}$ & \textbf{2.06} \\ \midrule
      JiT-H & 0.30 & 6 & $8.54\times10^{-3}$ & 1.85 \\
      EiT-H & \textbf{0.19} & \textbf{8} & $\mathbf{8.49\times10^{-3}}$ & \textbf{1.78} \\
      \bottomrule
   \end{tabular}
\end{table}

\section{Proofs for Orthonormal Transform Equivalence}
\label{Appendix:proof-orthonormal-equivalence}

\begin{proof}[Proof of Proposition~\ref{prop:orthonormal-equivalence}]
The explicit coupling $\pmb{\epsilon}_z=\pmb{W}\pmb{\epsilon}$ gives
\[
\pmb{z}_t
=(1-t)\pmb{W}\pmb{x}+t\pmb{W}\pmb{\epsilon}
=\pmb{W}\pmb{x}_t,
\]
and therefore
$\widetilde{\pmb{x}}_t=\pmb{W}^\top\pmb{z}_t=\pmb{x}_t$. The frequency-space
velocity target satisfies
\[
\pmb{v}_z
=\pmb{W}\pmb{\epsilon}-\pmb{W}\pmb{x}
=\pmb{W}\pmb{v}.
\]
Thus, for every realization of all coupled training randomness and every
$\theta$, orthonormality yields
\[
\begin{aligned}
L_{\mathrm{freq}}(\theta)
&=\|\pmb{W}\pmb{v}_\theta(\pmb{x}_t,t)
-\pmb{W}\pmb{v}\|_2^2 \\
&=\|\pmb{W}(\pmb{v}_\theta(\pmb{x}_t,t)-\pmb{v})\|_2^2 \\
&=\|\pmb{v}_\theta(\pmb{x}_t,t)-\pmb{v}\|_2^2
=L_{\mathrm{pix}}(\theta).
\end{aligned}
\]
Because the two sides are the same scalar function of $\theta$,
\[
\nabla_\theta L_{\mathrm{freq}}(\theta)
=
\nabla_\theta L_{\mathrm{pix}}(\theta)
\]
wherever differentiable. Averaging over a coupled minibatch preserves this
gradient identity, as does taking the expectation that defines the population
objectives.

Let $s_k^{\mathrm{pix}}$ and $s_k^{\mathrm{freq}}$ denote the optimizer
states. By assumption,
\[
(\theta_0^{\mathrm{pix}},s_0^{\mathrm{pix}})
=
(\theta_0^{\mathrm{freq}},s_0^{\mathrm{freq}}).
\]
Assume the two parameter--optimizer-state pairs are equal at step $k$. Since
all training randomness at this step is coupled, the two runs evaluate equal
minibatch gradients. Applying the same optimizer update to equal parameters,
states, and gradients gives
\[
(\theta_{k+1}^{\mathrm{pix}},s_{k+1}^{\mathrm{pix}})
=
(\theta_{k+1}^{\mathrm{freq}},s_{k+1}^{\mathrm{freq}}).
\]
Induction proves equality at every step. Any converged parameter vectors are
therefore identical.
\end{proof}

\begin{proof}[Proof of Corollary~\ref{cor:uncoupled-training-equivalence}]
Let $\Xi_k^{\mathrm{pix}}$ and $\Xi_k^{\mathrm{freq}}$ collect all stochastic
inputs at step $k$, including the minibatch, noise, timestep, conditioning
variables, data augmentation, and model randomness. The two collections are
sampled independently across runs but, by assumption, their corresponding
components follow the same joint sampling laws. For the frequency-space
noise, define $\pmb{\eta}=\pmb{W}^\top\pmb{\epsilon}_z$. Orthonormal
invariance gives
\[
\pmb{\eta}\sim\mathcal{N}(0,\pmb{W}^\top\pmb{I}\pmb{W})
=\mathcal{N}(0,\pmb{I}).
\]
After mapping the frequency-space quantities back to pixel space,
\[
\widetilde{\pmb{x}}_t=(1-t)\pmb{x}+t\pmb{\eta},
\qquad
\pmb{W}^\top\pmb{v}_z=\pmb{\eta}-\pmb{x}.
\]
Together with the matching laws of all other components of $\Xi_k$, this
shows that the complete transformed frequency-space minibatch has the same
distribution as the pixel-space minibatch. Norm preservation then implies,
for every fixed $\theta$,
\[
\bigl(L_{\mathrm{freq}}(\theta),
\nabla_\theta L_{\mathrm{freq}}(\theta)\bigr)
\overset{d}{=}
\bigl(L_{\mathrm{pix}}(\theta),
\nabla_\theta L_{\mathrm{pix}}(\theta)\bigr).
\]
The same identity holds for the corresponding stochastic minibatch losses
and gradients.

Let
$S_k^{\mathrm{pix}}=(\theta_k^{\mathrm{pix}},s_k^{\mathrm{pix}})$ and
$S_k^{\mathrm{freq}}=(\theta_k^{\mathrm{freq}},s_k^{\mathrm{freq}})$ denote
the parameter--optimizer states. The initial states are sampled from the same
distribution; identical deterministic initialization is a special case.
Because both runs use the same optimizer configuration, the result above
implies that, conditional on any fixed state, their next-state updates have
the same transition distribution. If
$S_k^{\mathrm{freq}}\overset{d}{=}S_k^{\mathrm{pix}}$, applying these
identical transition distributions gives
$S_{k+1}^{\mathrm{freq}}\overset{d}{=}S_{k+1}^{\mathrm{pix}}$. Induction from
the initial states therefore yields
\[
(\theta_k^{\mathrm{freq}},s_k^{\mathrm{freq}})
\overset{d}{=}
(\theta_k^{\mathrm{pix}},s_k^{\mathrm{pix}})
\quad\text{for every }k.
\]

Moreover, the same input--target distribution and norm preservation give
$\mathcal{L}_{\mathrm{freq}}(\theta)
=\mathcal{L}_{\mathrm{pix}}(\theta)$ for every fixed $\theta$. Combining
this pointwise identity with equality in distribution of the parameter
iterates yields
\[
\mathbb{E}_{\mathrm{train}}\!\left[
\mathcal{L}_{\mathrm{freq}}(\theta_k^{\mathrm{freq}})
\right]
=
\mathbb{E}_{\mathrm{train}}\!\left[
\mathcal{L}_{\mathrm{pix}}(\theta_k^{\mathrm{pix}})
\right].
\]
The statement for converged iterates follows whenever the terminal iterates
and corresponding expected losses are well defined. Since the stochastic
variables are not coupled across runs, this argument does not imply equality
of individual optimization trajectories.
\end{proof}

\section{Empirical Verification of Equivalence}
\label{Appendix: Empirical Verification of Equivalence}

To empirically verify the equivalence in Section~\ref{Subsection: Equivalence of Frequency-space Diffusion and Pixel-space Diffusion}, we train JiT-B in pixel space and JiT-B (DWT) in the Haar DWT space following Figure~\ref{fig: DWT Flow Matching}, without loss reweighting ($\lambda=\gamma=1$). The training randomness of the two runs is sampled independently, matching the uncoupled setting of Corollary~\ref{cor:uncoupled-training-equivalence}. As shown in Table~\ref{tab: Generation Comparison of Pixel Diffusion and Frequency Diffusion on ImageNet 256.}, the two models achieve nearly identical FID after 600 epochs on ImageNet 256$\times$256: 24.23 vs.\ 24.31 for conditional generation and 3.59 vs.\ 3.60 with CFG. Moreover, since the DWT is applied outside the network, JiT-B (DWT) keeps the same parameters and GFLOPs as JiT-B, incurring negligible extra training cost.

\begin{table}[ht]
   \centering
   \captionsetup{skip=2pt}
   \caption{Generation Comparison of Pixel Diffusion and Frequency Diffusion on ImageNet 256$\times$256.}
   \label{tab: Generation Comparison of Pixel Diffusion and Frequency Diffusion on ImageNet 256.}
      \begin{tabular}{@{}llllll@{}}
         \toprule
         \multirow{2}{*}{Models} & \multirow{2}{*}{Parameters} & \multirow{2}{*}{GFLOPs} & \multirow{2}{*}{Epochs} & \multicolumn{2}{c}{FID} \\ \cmidrule(l){5-6} 
          &  &  &  & cond & CFG \\ \midrule
         JiT-B & 131M & 25.2 & 600 & 24.23 & 3.59 \\
         JiT-B (DWT) & 131M & 25.2 & 600 & 24.31 & 3.60 \\
         JiT-B (vanilla DWT) & 131M & 25.2 & 600 & 25.11 & 3.71 \\ \bottomrule
         \end{tabular}
\end{table}

\paragraph{Feeding DWT coefficients directly to the network.} We further train JiT-B (vanilla DWT), which feeds the noisy DWT coefficients $\pmb{z}_t$ directly into the network, instead of first mapping them back to the pixel-space input $\widetilde{\pmb{x}}_t=\pmb{W}^\top\pmb{z}_t$ as in Figure~\ref{fig: DWT Flow Matching}. It is trained with the same DWT velocity loss ($\lambda=\gamma=1$), architecture, and training recipe as JiT-B (DWT), so the two models differ only in the input space of the network. As shown in Table~\ref{tab: Generation Comparison of Pixel Diffusion and Frequency Diffusion on ImageNet 256.}, JiT-B (vanilla DWT) degrades the FID from 24.31 to 25.11 for conditional generation and from 3.60 to 3.71 with CFG. This gap is consistent with our theory, since Proposition~\ref{prop:orthonormal-equivalence} requires the same pixel-space predictor to be applied to $\pmb{W}^\top\pmb{z}_t$. Once the network directly receives $\pmb{z}_t$, the equivalence no longer holds: the architecture is designed for the spatial structure of RGB images rather than for DWT coefficients, and adaptive optimizers such as Adam are not invariant to an orthonormal change of the input coordinates. In addition, the DWT sub-bands have very different dynamic ranges, with large LL coefficients and sparse, near-zero high-frequency coefficients, which may make training numerically more sensitive. In contrast, the design in Figure~\ref{fig: DWT Flow Matching} applies the DWT only outside the network and keeps the network operating in pixel space. It therefore inherits the inductive biases of pixel-space architectures, is exactly covered by Proposition~\ref{prop:orthonormal-equivalence}, and still enables frequency-band loss reweighting. We thus adopt this design for all frequency-space models in this paper.

\section{Training Parameters and Model Details}
\label{Appendix: Training Parameters and Model Details}

Table~\ref{tab: Configurations of experiments.} lists the configurations of JiT and EiT. EiT is built on JiT and shares its patch size, 128-dimensional bottleneck patch embedding, 32 in-context class tokens, training recipe, and sampler, so that the two models differ only in architecture.

\paragraph{Architecture.} Following Section~\ref{Subsection: U-shaped Architecture and Channel Compression}, EiT keeps the head dimension fixed ($d=64$ for EiT-B and $d=72$ for EiT-L) and compresses channels by reducing the number of heads. The u-type specifies the encoder stage widths in units of $D/4$, where $D$ is the maximum hidden dimension (channel dimension), and the decoder mirrors this schedule. Thus, $4$-$3$-$2$-$1$ in EiT-B corresponds to widths $(768, 576, 384, 192)$ with $(12, 9, 6, 3)$ heads, and its 24 blocks are evenly distributed over the eight encoder and decoder stages (3 blocks per stage). The hidden dimension and number of heads of EiT are marked with $\leq$ in the table. Adjacent stages are connected by linear projections, and mirrored encoder and decoder blocks by long skips. For all models, the in-context class tokens are inserted at one third of the depth for base and large models, and at one fourth of the depth for huge models.

\paragraph{Training and sampling.} Base and large models are trained for 600 epochs with Adam \citep{kingma2014adam}, a batch size of 2048, and a constant learning rate of $4\times10^{-4}$ after 5 warmup epochs. We keep two EMA models with decays 0.9999 and 0.9996, and drop class labels with probability 0.1 to enable CFG. We use patch sizes 16 and 32 for ImageNet 256$\times$256 and 512$\times$512, respectively, yielding the same 256 tokens, and increase the noise scale from 1.0 to 2.0 at the higher resolution. For sampling, we use a 50-step Heun solver with linearly spaced timesteps, apply CFG within the interval $t\in[0.1,1.0]$ \citep{kynkaanniemi2024applying}, and sweep the CFG scale over $[2.0, 4.0]$.

\begin{table}[ht]
   \vskip -0.0in
   \captionsetup{skip=2pt}
   \caption{
    Configurations of experiments.}
   \label{tab: Configurations of experiments.}
   \begin{center}
      \resizebox{\textwidth}{!}{
      \begin{tabular}{@{}lcl|llllll@{}}
         \toprule
          & \multicolumn{6}{c|}{ImageNet 256} & \multicolumn{2}{c}{ImageNet 512} \\ \cmidrule(l){2-9}
          & \multicolumn{1}{l}{\textbf{JiT-B}} & \textbf{EiT-B} & \textbf{JiT-L} & \multicolumn{1}{c|}{\textbf{EiT-L}} & \textbf{JiT-H} & \multicolumn{1}{c|}{\textbf{EiT-H}} & \textbf{JiT-H} & \multicolumn{1}{c}{\textbf{EiT-H}} \\ \midrule
         \rowcolor{gray!15} \multicolumn{9}{l}{\textbf{architecture}} \\
         depth & \multicolumn{1}{l}{12} & 24 & 24 & \multicolumn{1}{l|}{30} & 32 & \multicolumn{1}{l|}{40} & 32 & 40 \\
         hidden dim & \multicolumn{1}{l}{768} & $\leq$768 & 1024 & \multicolumn{1}{l|}{$\leq$1152} & 1344 & \multicolumn{1}{l|}{$\leq$1728} & 1344 & $\leq$1728 \\
         u-type & \multicolumn{1}{l}{-} & 4-3-2-1 & - & \multicolumn{1}{l|}{4-3-2} & - & \multicolumn{1}{l|}{4-3-2-1} & - & 4-3-2-1 \\
         heads & \multicolumn{1}{l}{12} & $\leq$12 & 16 & \multicolumn{1}{l|}{$\leq$16} & 16 & \multicolumn{1}{l|}{$\leq$16} & 16 & $\leq$16 \\
         parameters & \multicolumn{1}{l}{131M} & 130M & 459M & \multicolumn{1}{l|}{460M} & 1050M & \multicolumn{1}{l|}{1074M} & 1054M & 1079M \\
         patch size & \multicolumn{1}{l}{16} & 16 & 16 & \multicolumn{1}{l|}{16} & 16 & \multicolumn{1}{l|}{16} & 32 & 32 \\
         bottleneck dim & \multicolumn{1}{l}{128} & 128 & 128 & \multicolumn{1}{l|}{128} & 256 & \multicolumn{1}{l|}{256} & 256 & 256 \\
         dropout & \multicolumn{1}{l}{0} & 0 & 0 & \multicolumn{1}{l|}{0} & 0.2 & \multicolumn{1}{l|}{0.2} & 0.2 & 0.2 \\
         in-context class tokens & \multicolumn{1}{l}{32} & 32 & 32 & \multicolumn{1}{l|}{32} & 32 & \multicolumn{1}{l|}{32} & 32 & 32 \\
         in-context start block & \multicolumn{1}{l}{4} & 8 & 8 & \multicolumn{1}{l|}{10} & 8 & \multicolumn{1}{l|}{10} & 8 & 10 \\ \midrule
         \rowcolor{gray!15} \multicolumn{9}{l}{\textbf{training}} \\
         epochs & \multicolumn{2}{c|}{600} & \multicolumn{2}{c|}{600} & \multicolumn{2}{c|}{600} & \multicolumn{2}{c}{600} \\
         warmup epochs & \multicolumn{2}{c|}{5} & \multicolumn{2}{c|}{5} & \multicolumn{2}{c|}{5} & \multicolumn{2}{c}{5} \\
         optimizer & \multicolumn{2}{c|}{Adam} & \multicolumn{2}{c|}{Adam} & \multicolumn{2}{c|}{Adam} & \multicolumn{2}{c}{Adam} \\
         batch size & \multicolumn{2}{c|}{2048} & \multicolumn{2}{c|}{2048} & \multicolumn{2}{c|}{1024} & \multicolumn{2}{c}{1024} \\
         learning rate & \multicolumn{2}{c|}{4e-4} & \multicolumn{2}{c|}{4e-4} & \multicolumn{2}{c|}{2e-4} & \multicolumn{2}{c}{2e-4} \\
         learning rate schedule & \multicolumn{2}{c|}{constant} & \multicolumn{2}{c|}{constant} & \multicolumn{2}{c|}{constant} & \multicolumn{2}{c}{constant} \\
         ema decay & \multicolumn{2}{c|}{\{0.9999, 0.9996\}} & \multicolumn{2}{c|}{\{0.9999, 0.9996\}} & \multicolumn{2}{c|}{\{0.9999, 0.9996\}} & \multicolumn{2}{c}{\{0.9999, 0.9996\}} \\
         noise scale & \multicolumn{2}{c|}{1.0} & \multicolumn{2}{c|}{1.0} & \multicolumn{2}{c|}{1.0} & \multicolumn{2}{c}{2.0} \\
         class token drop & \multicolumn{2}{c|}{0.1} & \multicolumn{2}{c|}{0.1} & \multicolumn{2}{c|}{0.1} & \multicolumn{2}{c}{0.1} \\ \midrule
         \rowcolor{gray!15} \multicolumn{9}{l}{\textbf{sampling}} \\
         ODE solver & \multicolumn{2}{c|}{Heun} & \multicolumn{2}{c|}{Heun} & \multicolumn{2}{c|}{Heun} & \multicolumn{2}{c}{Heun} \\
         ODE steps & \multicolumn{2}{c|}{50} & \multicolumn{2}{c|}{50} & \multicolumn{2}{c|}{50} & \multicolumn{2}{c}{50} \\
         time steps & \multicolumn{2}{c|}{linear in {[}0.0, 1.0{]}} & \multicolumn{2}{c|}{linear in {[}0.0, 1.0{]}} & \multicolumn{2}{c|}{linear in {[}0.0, 1.0{]}} & \multicolumn{2}{c}{linear in {[}0.0, 1.0{]}} \\
         CFG scale sweep range & \multicolumn{2}{c|}{{[}2.0, 4.0{]}} & \multicolumn{2}{c|}{{[}2.0, 4.0{]}} & \multicolumn{2}{c|}{{[}2.0, 4.0{]}} & \multicolumn{2}{c}{{[}2.0, 4.0{]}} \\
         CFG interval & \multicolumn{2}{c|}{{[}0.1, 1.0{]}} & \multicolumn{2}{c|}{{[}0.1, 1.0{]}} & \multicolumn{2}{c|}{{[}0.1, 1.0{]}} & \multicolumn{2}{c}{{[}0.1, 1.0{]}} \\ \bottomrule
         \end{tabular}
      }
   \end{center}
   \vskip -0.0in
\end{table}

\section{Ablation Studies}
\label{Appendix: Ablation Studies}

\subsection{Effect of long-skip connection in JiT and EiT}
\label{Appendix: Effect of long-skip connection in JiT and EiT}

Table~\ref{tab: Ablation study of long-skip connections in JiT and EiT.} shows that long-skip connections have little effect on the plain transformer JiT-B (FID 3.59 without skips vs.\ 3.63 with skips), but substantially improve EiT-B (3.50 to 3.10). JiT keeps a constant hidden dimension across Transformer blocks, while EiT compresses its intermediate representations. As in UNet \citep{ronneberger2015u}, long skips provide an additional path around the bottleneck for high-frequency details that compression may discard. This explains why long-skip connections are important for EiT's U-shaped architecture.

\begin{table}[ht]
   \vskip -0.0in
   \captionsetup{skip=2pt}
   \caption{Ablation study of long-skip connections in JiT-B and EiT-B on ImageNet 256$\times$256.}
   \label{tab: Ablation study of long-skip connections in JiT and EiT.}
   \begin{center}
      \begin{tabular}{@{}lcllll@{}}
         \toprule
         Models & \shortstack{Long skips} & Parameters & GFLOPs & \shortstack[l]{Epochs} & FID (CFG) $\downarrow$ \\ \midrule
         JiT-B & $\times$ & 131M & 25.2 & 600 & \textbf{3.59} \\
         JiT-B & $\checkmark$ & 131M & 25.2 & 600 & 3.63 \\ \midrule
         EiT-B & $\times$ & 130M & 24.3 & 600 & 3.50 \\
         EiT-B & $\checkmark$ & 130M & 24.3 & 600 & \textbf{3.10} \\ \bottomrule
      \end{tabular}
   \end{center}
   \vskip -0.0in
\end{table}

\subsection{Design of U-type in EiT}
\label{Appendix: Design of U-type in EiT}

The u-type determines how EiT compresses channels and controls both the depth and the bottleneck width. U-type $4$-$3$-$2$-$1$ has eight stages with a bottleneck of $D/4$, whereas $4$-$3$-$2$ has six stages with a wider bottleneck of $D/2$ (where $D$ is the channel dimension). For example, EiT-B has 24 blocks with $4$-$3$-$2$-$1$ and 18 blocks with $4$-$3$-$2$. Adding a narrower stage therefore deepens the network at the cost of a narrower bottleneck.

Table~\ref{tab: Ablation study of U-type design in EiT.} compares the two schedules, and \textbf{all EiT variants outperform their JiT counterparts}. For the Base models, $4$-$3$-$2$-$1$ (24 blocks) achieves better generation quality than $4$-$3$-$2$ (18 blocks) by showing FID 3.10 vs.\ 3.34. For the Large models, U-type $4$-$3$-$2$-$1$ shows very close performance with u-type $4$-$3$-$2$. At the same maximum channel dimension of 1024, $4$-$3$-$2$ achieves similar results to $4$-$3$-$2$-$1$ in FID (2.22 vs.\ 2.19). Increasing the maximum channel dimension to 1152 benifits both u-tyeps in generation quality.

\begin{table}[ht]
   \vskip -0.0in
   \captionsetup{skip=2pt}
   \caption{
   Ablation study of the u-type design in EiT on ImageNet 256$\times$256.}
   \label{tab: Ablation study of U-type design in EiT.}
   \begin{center}
      \begin{tabular}{@{}llllllll@{}}
         \toprule
          & U-type & Depth & Hidden dim & Parameters & GFLOPs & Epochs & FID (CFG) \\ \midrule
         JiT-B & - & 12 & 768 & 131M & 25.2 & 600 & 3.59 \\
         EiT-B & 4-3-2-1 & 24 & $\leq$768 & 130M & 24.3 & 600 & \textbf{3.10} \\
         EiT-B & 4-3-2 & 18 & $\leq$768 & 125M & 23.3 & 600 & 3.34 \\ \midrule
         JiT-L & - & 24 & 1024 & 459M & 87.9 & 600 & 2.41 \\
         EiT-L & 4-3-2-1 & 40 & $\leq$1024 & 378M & 69.8 & 600 & 2.19 \\
         EiT-L & 4-3-2 & 30 & $\leq$1024 & 363M & 67.1 & 600 & 2.22 \\
         EiT-L & 4-3-2-1 & 40 & $\leq$1152 & 478M & 87.8 & 600 & 2.09 \\
         EiT-L & 4-3-2 & 30 & $\leq$1152 & 460M & 84.4 & 600 & \textbf{2.06} \\ \bottomrule
         \end{tabular}
   \end{center}
   \vskip -0.0in
\end{table}

\subsection{Contribution of Channel Compression in EiT}
\label{Appendix
: Contribution of Channel Compression in EiT}

In Section \ref{subection: results of channel compression - EiT} we show that EiT consistently outperforms JiT with a difference layer depth. To validate the 
contribution of channel compression in EiT, we conduct the abation study by fixing the Transformer depth at 24 for all models.
Results in Table~\ref{tab: Ablation study of channel compression in EiT.} support channel compression as a key contributor to EiT's performance gains even when network depth is fixed.

\begin{table}[ht]
   \vskip -0.0in
   \captionsetup{skip=2pt}
   \caption{
   Ablation study of the contribution of channel compression in EiT on ImageNet 256$\times$256.}
   \label{tab: Ablation study of channel compression in EiT.}
   \begin{center}
      \begin{tabular}{@{}llllllll@{}}
         \toprule
          & U-type & Depth & Hidden dim & Parameters & GFLOPs & Epochs & FID (CFG) \\ \midrule
         JiT-L & - & 24 & 1024 & 459M & 87.9 & 600 & 2.41 \\
         PixelU-L & - & 24 & 1024 & 516M & 75.3 & 600 & 2.23 \\
         EiT-L & 4-3-2 & 24 & $\leq$1344 & 469M & 87.8 & 600 & \textbf{2.11} \\ \bottomrule
         \end{tabular}
   \end{center}
   \vskip -0.0in
\end{table}

\subsection{Loss Reweighting of DWT and Low-Resolution Regularization}
\label{Appendix: Loss Reweighting of DWT and Low-Resolution Regularization}

Section~\ref{Subsection: Equivalence of Frequency-space Diffusion and Pixel-space Diffusion} introduces two ways to strengthen the modeling of low-frequency content: DWT loss reweighting (Equation~\ref{eq: dwt-velocity-loss}) and Low-Resolution Regularization (LRR, Equation~\ref{eq: low-resolution-regularization}). Table~\ref{tab: Ablation study of DWT and Low-Resolution Regularization in EiT.} compares the two methods on EiT-B ($4$-$3$-$2$-$1$) on ImageNet 256$\times$256, where all models are trained for 600 epochs with the same recipe, and EiT-B (DWT) uses different $\lambda$ and $\gamma=1$. For EiT-B (LRR), the low-resolution image $\pmb{x}_{\mathrm{low}}$ is obtained by bicubic downsampling, and the weight of the low-resolution loss in Equation~\ref{eq: low-resolution-regularization} is set to $\eta=1.2$. By Proposition~\ref{prop:orthonormal-equivalence}, the pixel-space EiT-B is equivalent to its DWT counterpart with $\lambda=\gamma=1$, so the comparison between EiT-B and EiT-B (DWT) isolates the effect of up-weighting the LL band.

\paragraph{Similar gains.} Both methods improve the CFG FID of EiT-B by a similar margin: DWT loss reweighting reduces it from 3.10 to 3.02, and LRR reduces it to 3.00. This agreement suggests that the two methods benefit from the same source, namely emphasizing low-frequency content during training, which is beneficial for generation on ImageNet.

\paragraph{Different costs.} The two methods differ substantially in training cost. Since the orthonormal DWT is applied outside the network, EiT-B (DWT) keeps exactly the same parameters (130M) and GFLOPs (24.3) as EiT-B. In contrast, LRR requires an additional forward pass on the low-resolution input at every training iteration, which raises the training cost from 24.3 to 31.0 GFLOPs (about 28\% more) while yielding a nearly identical FID (3.00 vs.\ 3.02). We therefore adopt DWT loss reweighting as the default low-frequency enhancement in our main experiments, as it achieves comparable improvements at no extra cost.

\begin{table}[ht]
   \vskip -0.0in
   \captionsetup{skip=2pt}
   \caption{
   Ablation study of the DWT loss reweighting and Low-Resolution Regularization in EiT on ImageNet 256$\times$256.}
   \label{tab: Ablation study of DWT and Low-Resolution Regularization in EiT.}
   \begin{center}
      \begin{tabular}{@{}lllllll@{}}
         \toprule
          & U-type & Depth & Hidden dim & Parameters & GFLOPs & FID (CFG) \\ \midrule
         EiT-B & 4-3-2-1 & 24 & $\leq$768 & 130M & 24.3 & 3.10 \\
         EiT-B (DWT, $\lambda=1.1$) & 4-3-2-1 & 24 & $\leq$768 & 130M & 24.3 & 3.05 \\
         EiT-B (DWT, $\lambda=1.2$) & 4-3-2-1 & 24 & $\leq$768 & 130M & 24.3 & \textbf{3.02}  \\
         EiT-B (DWT, $\lambda=1.3$) & 4-3-2-1 & 24 & $\leq$768 & 130M & 24.3 & 3.07 \\
         EiT-B (LRR, $\eta=1.2$) & 4-3-2-1 & 24 & $\leq$768 & 131M & 31.0 & \textbf{3.00} \\ \bottomrule
         \end{tabular}
   \end{center}
   \vskip -0.0in
\end{table}

\end{document}